\documentclass[conference]{IEEEtran}
\IEEEoverridecommandlockouts

\usepackage{cite}
\usepackage{amsmath,amssymb,amsfonts}
\usepackage{algorithmic}
\usepackage{graphicx}
\usepackage{textcomp}
\usepackage{xcolor}
\usepackage{url} 
\usepackage{hyperref}

\def\BibTeX{{\rm B\kern-.05em{\sc i\kern-.025em b}\kern-.08em
  T\kern-.1667em\lower.7ex\hbox{E}\kern-.125emX}}
\begin{document}

\title{DMV-AVP: Distributed Multi-Vehicle Autonomous Valet Parking Using Autoware}

\author{
\IEEEauthorblockN{Zubair Islam}
\IEEEauthorblockA{
\textit{Faculty of Engineering and Applied Science} \\
\textit{Ontario Tech University}\\
Oshawa, Canada \\
zubair.islam@ontariotechu.net}
\and
\IEEEauthorblockN{Mohamed El-Darieby}
\IEEEauthorblockA{
\textit{Faculty of Engineering and Applied Science} \\
\textit{Ontario Tech University}\\
Oshawa, Canada \\
Mohamed.El-Darieby@ontariotechu.ca}
}
\maketitle

\begin{abstract}
This paper presents DMV-AVP, a distributed simulation of Multi-Vehicle Autonomous Valet Parking (AVP). The system was implemented as an application of the Distributed Multi-Autonomous Vehicle Architecture (DMAVA) for synchronized multi-host execution. Most existing simulation approaches rely on centralized or non-distributed designs that constrain scalability and limit fully autonomous control. This work introduces two modules built on top of DMAVA: 1) the Multi-Vehicle AVP Coordination Framework, composed of AVP Managers and a per-vehicle AVP Node, is responsible for global parking state tracking, vehicle queuing, parking spot reservation, lifecycle coordination, and conflict resolution across multiple vehicles, and 2) the Unity-Integrated YOLOv5 Parking Spot Detection Module, that provides real-time, vision-based perception within AWSIM Labs. Both modules integrate seamlessly with DMAVA and extend it specifically for multi-vehicle AVP operation, supported by a Zenoh communication layer that ensures high data accuracy and controllability across hosts. Experiments conducted on two- and three-host configurations demonstrate consistent coordination, conflict-free parking behavior, and scalable performance across distributed Autoware instances. The results confirm that the proposed DMV-AVP supports cooperative AVP simulation and establishes a foundation for future real-world and hardware-in-the-loop validation. Demo videos and source code are available at \href{https://github.com/zubxxr/multi-vehicle-avp}{\textbf{https://github.com/zubxxr/multi-vehicle-avp}}

\end{abstract}


\section{Introduction}
\label{introduction}
Multi-vehicle autonomous driving simulations are essential for evaluating realistic interactions among autonomous vehicles (AVs). Simulation environments play a critical role in balancing accuracy, scalability, and cost during the validation and verification of AV systems. Multi-AV operation in close proximity represents a particularly challenging problem, as it introduces constraints related to spatial coordination and communication synchronization. Autonomous Valet Parking (AVP) exemplifies this problem in structured, low-speed environments. Unlike open-road driving, parking environments are dense and require fine-grained localization, cooperative behavior, and efficient multi-AV coordination. Simulation provides a reproducible platform for investigating these challenges prior to real-world deployment.

Research into AVP has explored both onboard and infrastructure-supported approaches. TIER IV and the Autoware Foundation demonstrated an onboard-only single-vehicle AVP system using the LGSVL simulator \cite{b1} and later extending the setup to a real-world deployment \cite{b2} with LiDAR and cameras. However, LGSVL was deprecated in 2022, and the AVP implementation relied on an older Autoware version built on the legacy ROS 2 Foxy distribution. Subsequent simulation-based AVP efforts using the CARLA simulator have remained limited to single-vehicle operation, relying on predefined trajectories without adaptive perception or vehicle-to-vehicle (V2V) coordination \cite{b3}. Infrastructure-based systems, such as Bosch and Mercedes deployments at Stuttgart Airport \cite{b4} and Enterprise’s commercial testing of Bosch AVP technology \cite{b5}, demonstrate the feasibility of infrastructure-assisted parking but do not address decentralized or onboard multi-vehicle autonomy.

To address the need for distributed multi-AV operation, this work builds upon the architectural foundations of the Distributed Multi-AV Architecture (DMAVA) \cite{b6}. DMAVA integrates ROS 2 Humble, Autoware Universe \cite{b7} for autonomous driving, AWSIM Labs \cite{b8} for simulation, and Zenoh \cite{b9} for inter-host communication. While DMAVA provides the core infrastructure, it lacks application-level coordination logic, with vehicle operation remaining user-driven and requiring manual goal pose specification and explicit activation of AD for each vehicle.

This paper presents Distributed Multi-Vehicle AVP (DMV-AVP), which extends DMAVA to enable coordinated AVP and address key requirements for realistic multi-AV parking scenarios. The proposed system introduces three novel extensions: (i) autonomous goal assignment and coordination through the Multi-Vehicle AVP Coordination Framework (AVP-CF), composed of centralized AVP Managers and per-vehicle AVP Nodes that implement state-based coordination, vehicle queuing, and parking-spot reservation, enabling vehicles to autonomously assign and manage navigation goals throughout the AVP lifecycle after high-level user commands such as drop-off or park, (ii) conflict-free reservation logic that maintains consistent parking-spot allocation and synchronized vehicle states across distributed Autoware instances, and (iii) infrastructure-based perception integration within AWSIM Labs through the Unity-Integrated YOLOv5 Parking Spot Detection (U-YOLO) Module, which provides real-time, vision-based parking availability directly within the simulation environment and enables shared infrastructure-assisted perception across all vehicles.

The remainder of this paper is organized as follows. Section~\ref{related_works} reviews related work on AVP systems and multi-AV coordination. Section~\ref{system_design} describes the system architecture and implementation of DMV-AVP. Section~\ref{experiments_results} describes the experimental setup and results obtained from two- and three-host configurations, validating system scalability and performance, while Section~\ref{discussion_conclusions} concludes the paper, discussing strengths, limitations, and future work.

\section{Related Works}
\label{related_works}
Research on AVP primarily focuses on distributed multi-AV coordination and vision-based parking-spot detection. Existing solutions typically rely on centralized control, single-host execution, or perception pipelines that do not support real-time coordination across multiple AVs. This section reviews prior work in multi-AV coordination architectures and infrastructure-assisted vision systems.

Several works address cooperative planning and control. Kneissl et al. \cite{b10, b11} proposed multi-vehicle AVP systems using the Vires VTD simulator with centralized coordination, extending their initial work from single-host to distributed components but remaining confined to ROS 1 and VTD environments. Liu et al. \cite{b12} introduced a Bio-Inspired Evolutionary Reinforcement Learning architecture for multi-AV collaborative trajectory planning in parking-lot scenarios, demonstrating strong algorithmic performance but remaining purely simulation-based without real-time autonomy or distributed operation.

Within the Autoware ecosystem, multi-AV coordination was first demonstrated with CARLA \cite{b13}, establishing a foundational architecture for running multiple Autoware instances using Zenoh for topic routing and namespace isolation. The Traffic Management System prototype \cite{b14} extended this to coordinate traffic-light control for emergency vehicles. However, both implementations relied on centralized coordination and were confined to single-host execution.

These studies established the foundation for distributed Autoware communication using Zenoh, which DMV-AVP extends within AWSIM Labs for scalable and decentralized coordinated AVP across multiple hosts.

Vision-based parking spot detection has been widely adopted in AVP systems and is generally treated as a mature enabling technology rather than a primary research contribution. Prior work has explored both onboard approaches, which rely on surround-view or fisheye camera setups for parking-slot recognition \cite{b15, b16}, and infrastructure-based approaches that offload perception to external cameras for centralized occupancy monitoring \cite{b17, b18, b19}. While onboard methods eliminate the need for fixed infrastructure, they often require multiple cameras, careful calibration, and higher computational overhead. Infrastructure-based perception offers a practical alternative by providing a global view of the parking environment and simplifying vehicle-side requirements. 

Consistent with this paradigm, the proposed system integrates an overhead camera and YOLOv5-based detection directly within AWSIM Labs to provide real-time parking availability shared across all vehicles through ROS 2 and Zenoh, enabling synchronized multi-AV coordination without introducing additional onboard perception complexity.

\section{System Design and Architecture}
\label{system_design}
This section presents the design of DMV-AVP, extending DMAVA to support distributed coordination among multiple AVs.

\subsection{Extension of DMAVA}
DMAVA provides five foundational workflows for distributed multi-AV operation: Mapping, Simulation, Communication, Localization, and Autonomous Driving, as explained in detail in \cite{b6}. These workflows define the functional responsibilities and data flow within the architecture and are realized as containerized execution units. The Communication workflow is implemented through embedded Zenoh middleware rather than as a standalone container, with Zenoh instances operating within each functional container.

DMV-AVP extends this foundation by introducing two additional workflows tailored to AVP scenarios, which are likewise realized as containerized execution units. The Perception Workflow augments the Simulation Workflow by extracting parking availability from simulated sensor data through the U-YOLO Module and publishing this information into the system. The Coordination Workflow augments the Autonomous Driving Workflow by managing vehicle state, goal assignment, drop-off queuing, and parking spot reservation through the AVP-CF. These workflows interact with existing DMAVA components to influence autonomous execution while preserving the underlying planning and control architecture.

\begin{figure*}[t]
  \centering
  \includegraphics[width=\textwidth]{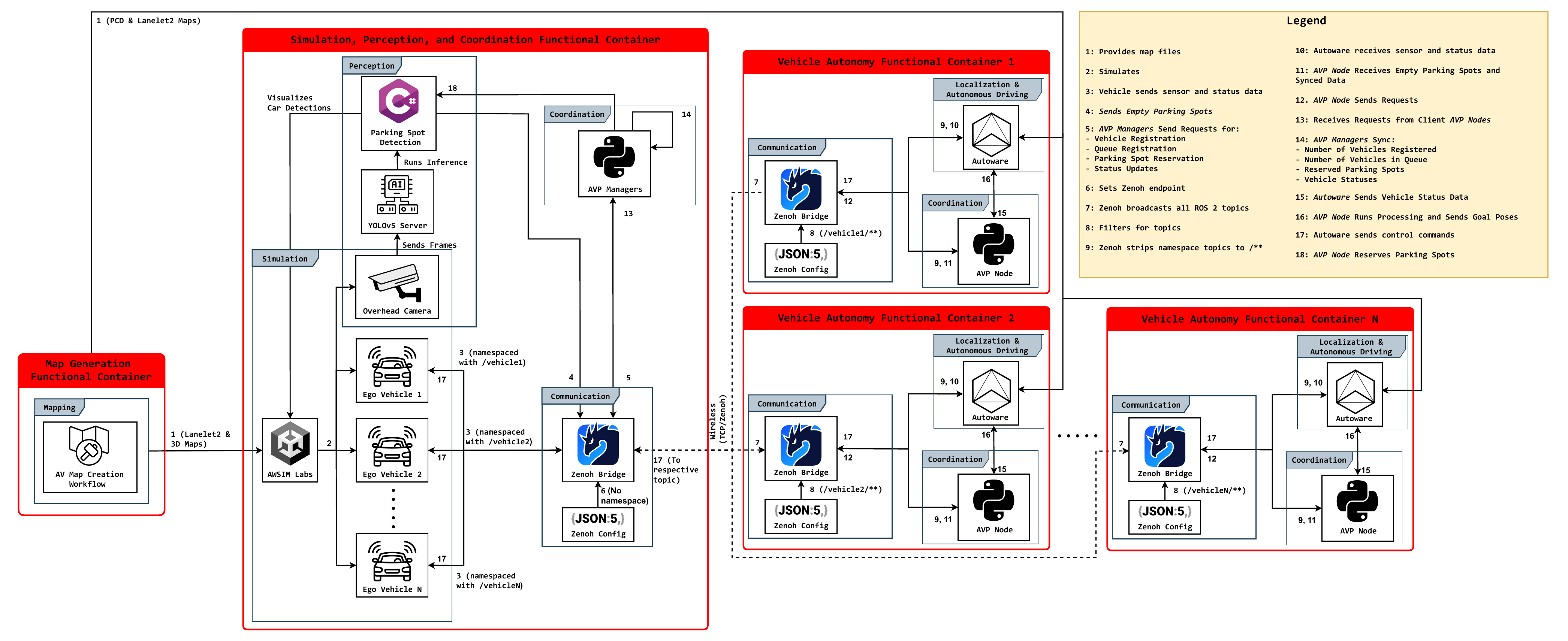}
  \caption{System-level container-based architecture of DMV-AVP.}
  \label{system_workflow}
\end{figure*}

Figure~\ref{system_workflow} illustrates the resulting DMV-AVP system organization, in which related workflow containers are grouped into higher-level functional containers based on their execution role, including Simulation, Perception, and Coordination, and Vehicle Autonomy. 

The Simulation, Perception, and Coordination functional container runs AWSIM Labs which simulates a shared parking environment containing multiple ego vehicles, each operating under unique ROS 2 namespaces. An overhead infrastructure camera, acting as a Roadside Unit (RSU), provides a top-down view of the parking area, with image frames streamed to the U-YOLO Module and processed to generate parking spot availability information shared with all vehicles. Due to infrastructure-assisted perception, fully decentralized V2V-only perception is outside the scope of this work.

Each Vehicle Autonomy functional container runs an independent Autoware instance responsible for per-vehicle localization, planning, and control. An embedded AVP Node within each container interfaces with the AVP Managers in the AVP-CF, enabling coordinated AVP execution across distributed vehicles. Zenoh selectively exposes and remaps ROS 2 topics between all containers, enabling distributed execution without modifying Autoware’s internal configuration, which preserves isolation while enabling coordinated operation within a shared simulated environment. It is important to note that, in this context, distributed refers to the execution of multiple independent Vehicle Autonomy functional containers across separate computing instances, while coordination decisions are derived from shared state maintained by the AVP Managers rather than through fully decentralized V2V decision-making.

\subsection{U-YOLO Module}
Reliable identification of parking spot availability is a prerequisite for coordinated multi-vehicle AVP operation, particularly in infrastructure-assisted scenarios where a shared global view must be maintained across distributed hosts. To address this requirement, DMV-AVP introduces the U-YOLO Module, which provides real-time detection of occupied and vacant parking spaces using an overhead RSU-mounted camera.

The module captures camera frames and forwards them to an external inference server, where vehicle detections are produced and returned to AWSIM Labs. Custom Unity C\# scripts then transform these detections into the simulated world frame and associate detected vehicle bounding boxes with predefined rectangular parking regions, each assigned a unique identifier. Parking spot availability is determined through geometric overlap checks, with overlapping regions marked as occupied and non-overlapping regions marked as available. The resulting list of available spot IDs is published as ROS 2 topics under vehicle-specific namespaces and synchronized across hosts via Zenoh, where the AVP-CF enforces a consistent, globally aligned view of parking availability through its manager components, and distributes this state to individual AVP Nodes. As the U-YOLO Module relies on a publicly available pre-trained YOLOv5 model \cite{b20} without additional fine-tuning, this work emphasizes distributed coordination, synchronization, and system-level integration rather than detector performance. Implementation details and source code are provided in \cite{b21}.

\subsection{Multi-Vehicle AVP Coordination Framework}
\begin{figure*}[t]
  \centering
  \includegraphics[width=\textwidth]{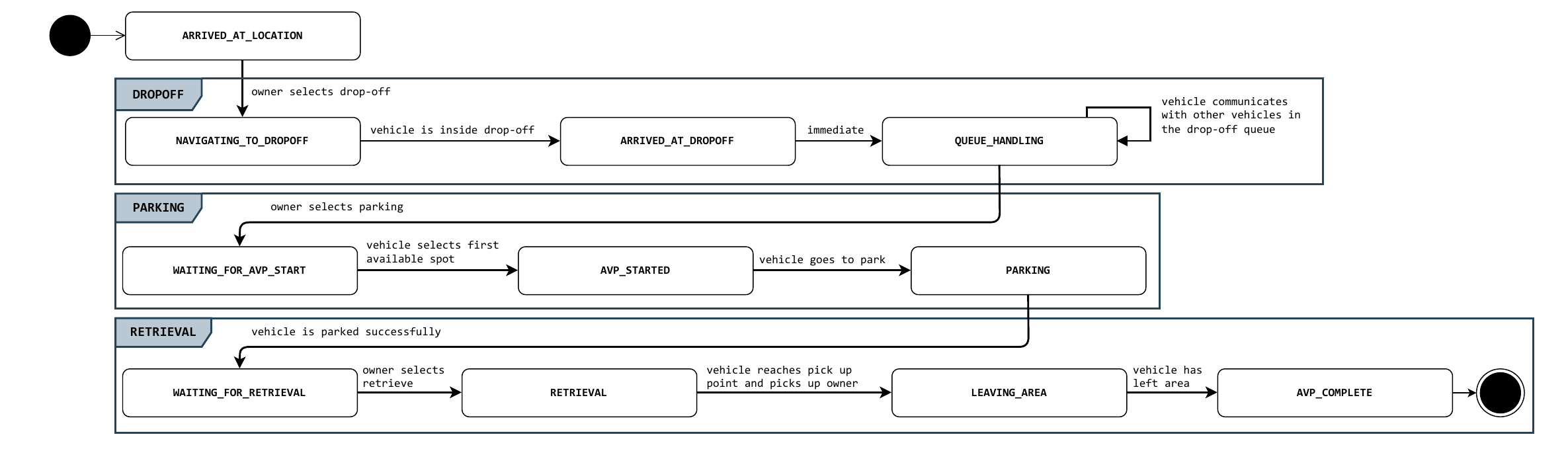}
  \caption{AVP state machine diagram illustrating the vehicle lifecycle from arrival to completion across Drop-off, Parking, and Retrieval phases.}
  \label{state_machine_diagram}
\end{figure*}

The AVP-CF serves as the core coordination framework enabling multiple AVs to execute drop-off, parking, and retrieval operations in a synchronized and conflict-free manner. The AVP-CF is decomposed into two complementary roles: the AVP Managers and a per-vehicle AVP Node. The AVP Managers are hosted in the Simulation, Perception, and Coordination functional container and are responsible for coordination logic spanning all vehicles. Rather than relying on direct V2V communication, coordination is achieved through shared system state maintained and propagated by these managers. The manager set consists of four components: a Vehicle Count Manager for tracking active vehicles, a Status Manager for publishing and synchronizing vehicle lifecycle states, a Queue Manager for ordering vehicles within the drop-off zone, and a Reservation Manager for managing parking spot allocation and preventing conflicts. The AVP Node is responsible for executing vehicle-specific AVP logic based on coordination decisions received from the AVP Managers. It interfaces with Autoware to issue goals, monitor execution status, and report state transitions back to the managers. Its behavior is governed by an event-driven state machine, illustrated in Figure~\ref{state_machine_diagram}, that spans the complete AVP lifecycle across the Drop-off, Parking, and Retrieval phases. State transitions are driven by vehicle status updates, coordination signals, and user-issued commands. 

A custom RViz-based AVP Panel was developed to support system visualization and manual control during testing. The panel provides real-time visibility into vehicle state, such as active vehicle count, per-vehicle lifecycle status, drop-off zone queue order, and available and reserved parking spots. It also exposes user controls for initiating Drop-off, Parking, and Retrieval actions, enabling streamlined testing and debugging of the multi-vehicle AVP lifecycle across distributed hosts. The complete implementation of the AVP-CF, including all AVP Managers and AVP Node source code, is available in \cite{b21}. 

\section{Experiments and Results}
\label{experiments_results}
Building on the previously validated DMAVA baseline, this section evaluates DMV-AVP under distributed execution.

\subsection{Experimental Setup}

Experiments were conducted using the same hardware configuration as reported in DMAVA \cite{b6}, consisting of two hosts with 24 GB of RAM and one host with 16 GB of RAM, each equipped with a discrete NVIDIA GPU. Scalability and system performance were evaluated using two distributed configurations: a primary two-host setup based on the first two hosts described in DMAVA, and an extended three-host setup introduced to assess behavior under increased computational load.

All experiments reused the DMAVA software stack \cite{b6}, with container placement following the DMAVA baseline configuration and extended with the AVP-CF and U-YOLO modules. In the two-host configuration, the Simulation, Perception, and Coordination functional container and one Vehicle Autonomy functional container were co-located on Host 1, while a second Vehicle Autonomy functional container was deployed on the Host 2. The three-host configuration extended this deployment by assigning an additional Vehicle Autonomy functional container to Host 3. This co-location was performed solely for practical resource management and does not represent a limitation of DMV-AVP. System startup followed a defined sequence to ensure consistent synchronization across all hosts and consistent topic alignment. Host 1 launched the simulation in AWSIM Labs, followed by the U-YOLO Module and its Autoware instance. Host 2 then launched its own Autoware instance. Once both Autoware stacks were active, Zenoh was executed on each host to establish inter-host communication. Finally, the AVP Managers are launched on Host 1 to initialize global coordination state, followed by the per-vehicle AVP Nodes on the secondary hosts. This startup procedure ensures repeatable initialization and consistent topic discovery during experimentation.

The RSU and AVP Managers are executed on the same physical host for deployment simplicity and have no process-level coupling. These components are implemented as logically independent modules that communicate exclusively through ROS 2 topics bridged via Zenoh. As such, architectural separation between vehicle autonomy, infrastructure-based perception, and coordination logic is preserved. 

\subsection{Distributed Validation and Scalability Testing}
\label{distributed_validation_scalability_testing}

DMV-AVP demonstrated stable two-host operation, with both vehicles successfully performing autonomous drop-off and parking without collisions. Vehicle status messages remained synchronized, and queue management maintained consistent scheduling, confirming coordinated multi-AV execution. The U-YOLO Module identified occupied and available parking spaces in real time, and Zenoh ensured stable topic synchronization across all components. 

\begin{figure}[h]
  \centering
  \includegraphics[width=0.9\linewidth]{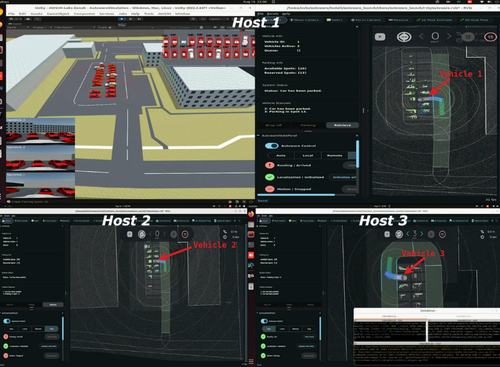}
  \caption{Three-host validation showing synchronized AVP operation across Hosts 1, 2, and 3, with all vehicles parked and system states aligned.}
  \label{multi_host}
\end{figure}

However, the retrieval phase was not functional due to limitations with Autoware’s Freespace Planner, and detection robustness was sensitive to vehicle appearance, which are both discussed further in Section~\ref{discussion_conclusions}. When extending from two to three vehicles, autonomous operation was maintained across all hosts. All vehicles received navigation goals, generated valid trajectories, and executed closed-loop planning and control without manual intervention. Results for both host configurations are included in the consolidated visualization in Figure~\ref{multi_host}.

During prolonged three-host execution, intermittent instability was observed. Although these effects did not prevent task completion or coordinated AVP behavior, they motivated further investigation into system behavior under increased load. The system-level performance characteristics and contributing factors associated with these observations are discussed in the following section. Video demonstrations of both two-host and three-host configurations are available in \cite{b21}.

\subsection{System Performance and Communication Timing}
\label{sec:system_performance}

System performance during active multi-vehicle AVP execution was evaluated under both two-host and three-host configurations. Consistent with DMAVA \cite{b6}, intermittent instability was observed during prolonged three-host operation. Analysis indicated that neither CPU utilization nor memory availability were primary contributors to the observed behavior, and that nominal inter-host communication latency alone does not fully explain the instability. Importantly, these effects did not prevent task completion or coordinated AVP behavior. The addition of the U-YOLO Module and the AVP-CF introduces increased computational and communication demands, motivating further quantitative evaluation of memory availability, CPU utilization, and inter-host communication latency.

Across all configurations, available memory remained high (Table~\ref{tab:memory_after_launch}), and CPU utilization remained within stable operating limits (Table~\ref{tab:cpu_utilization}), with both values higher than those reported in DMAVA. RTT measurements (Table~\ref{tab:rtt_three_host}) are reported as mean ± standard deviation, where the standard deviation reflects latency jitter, and were collected using the dedicated access point setup introduced in DMAVA. Baseline and idle RTT characteristics were previously evaluated in DMAVA. Therefore, this analysis focuses on RTT behavior during active AVP operation.

\begin{table}[t]
\centering
\caption{Available System Memory for Two- and Three-Host Configurations}
\label{tab:memory_after_launch}
\begin{tabular}{|c|c|c|}
\hline
\textbf{Setup} & \textbf{Host} & \textbf{Available Memory} \\
\hline
Two-Host & ROG Laptop & 6.6 GiB \\
     & Nitro PC  & 9.4 GiB \\
\hline
Three-Host & ROG Laptop  & 6.0 GiB \\
      & Nitro PC   & 9.4 GiB \\
      & Victus Laptop & 5.0 GiB \\
\hline
\end{tabular}
\end{table}

\begin{table}[t]
\centering
\caption{CPU Utilization for Two-Host and Three-Host Configurations}
\label{tab:cpu_utilization}
\begin{tabular}{|c|c|c|c|}
\hline
\textbf{Configuration} & \textbf{Host} & \textbf{Mean CPU (\%)} & \textbf{Peak CPU (\%)} \\
\hline
Two-Host  & ROG Laptop  & 62 & $<95$ \\
      & Nitro PC   & 48 & $<60$ \\
\hline
Three-Host & ROG Laptop  & 72 & $<95$ \\
      & Nitro PC   & 54 & $<65$ \\
      & Victus Laptop & 46 & $<75$ \\
\hline
\end{tabular}
\end{table}

\begin{table}[t]
\centering
\caption{RTT Measurements During Three-Host Active Operation}
\label{tab:rtt_three_host}
\begin{tabular}{|c|c|c|c|}
\hline
\textbf{Configuration} & \textbf{RTT (ms)} & \textbf{Max RTT (ms)} & \textbf{Samples} \\
\hline
Two-Host & $26.06 \pm 14.53$ & 100.74 & 2019 \\
\hline
Three-Host (ROG) & $16.24 \pm 15.02$ & 93.89 & 1976 \\
\hline
Three-Host (Victus) & $30.42 \pm 17.53$ & 138.11 & 1965 \\
\hline
\end{tabular}
\end{table}

The observed RTT values are lower than those reported in DMAVA, as measurements were collected on different days and are subject to time-varying wireless network conditions, including background interference and channel contention. As a result, direct comparison of absolute RTT magnitudes across studies should be interpreted with caution. Nevertheless, consistent with DMAVA, no sustained packet loss was observed during active AVP execution, and observed timing fluctuations were more strongly associated with network variability than with CPU utilization or memory availability.

\section{Discussion and Conclusions}
\label{discussion_conclusions}

DMV-AVP represents, to the authors’ knowledge, the first distributed multi-vehicle AVP simulation architecture that integrates Autoware Universe, AWSIM Labs, Zenoh, and infrastructure-based vision within a unified and fully open-source environment. Building upon the DMAVA, this work extends distributed autonomy to coordinated AVP by introducing the U-YOLO Module and the AVP-CF, enabling synchronized drop-off, parking, and reservation management across multiple AVs.

The evaluation of DMV-AVP using two- and three-host configurations demonstrates feasibility and correctness under constrained, household-grade computing resources. While the scale of the experimental setup does not constitute large-scale system scalability, these results validate the architectural foundations of DMAVA and DMV-AVP, including namespace isolation, inter-host synchronization, and coordinated multi-AV operation. Observed instability during prolonged three-host execution, consistent with prior DMAVA evaluations, was not attributable to sustained CPU or memory exhaustion. Instead, RTT analysis and system observations indicate that communication stability and timing variability under concurrent multi-host operation were the dominant contributing factors at this scale.

The RSU is co-located with simulation for deployment simplicity. However, the RSU logic is implemented as an independent component and can be isolated on separate infrastructure in realistic deployments, communicating with vehicles via Zenoh or other ROS 2–compatible transport mechanisms. The system further assumes that the RSU provides sufficient camera coverage of the parking area. Factors such as occlusion, limited field of view, calibration drift, and RSU failure modes are not explicitly modeled in the current simulation setup. In addition, detection performance exhibited sensitivity to vehicle appearance, as the system relied on a pre-trained YOLOv5 model without additional fine-tuning. As a result, parked vehicles in simulation were standardized to a single sedan model to maintain consistent occupancy classification. While these assumptions are reasonable for controlled AVP environments such as structured parking facilities, they limit the ability of the current evaluation to capture degraded or partially observable perception conditions.

The current implementation employs a centralized coordination model in which all decision-making is handled by the AVP Managers running on a single host. This design simplifies global state management, coordination logic, and synchronization across vehicles, but it also introduces a single coordination point that limits fault tolerance and redundancy. While suitable for controlled simulation-based validation, this architectural choice constrains resilience under failures or communication disruption.

Vehicle retrieval remains a known limitation, as Autoware’s Freespace Planner was unable to generate valid trajectories for vehicles parked outside Lanelet-covered regions, resulting in failed retrieval attempts. Although a manual workaround was implemented using Autoware’s Planning Simulator to reposition vehicles onto valid Lanelets, this approach bypasses normal operation and is unsuitable for production use. 

Planned future extensions therefore include enhancing the U-YOLO Module to achieve robust detection across diverse vehicle types and visual conditions, and incorporating more realistic infrastructure perception models that account for partial coverage, occlusion, calibration uncertainty, and RSU failure modes. Additional future work includes deploying the system on real robotic platforms for hardware-in-the-loop validation and exploring decentralized or fault-tolerant coordination strategies, in which the AVP Managers are distributed across multiple hosts to reduce reliance on a single coordination point.

\vspace{12pt}

\end{document}